\documentclass[10pt,conference]{IEEEtran}
\IEEEoverridecommandlockouts

\usepackage[T1]{fontenc}
\usepackage[utf8]{inputenc}
\usepackage{amsmath,amssymb,amsfonts}
\usepackage{amsthm}
\usepackage{mathtools}
\usepackage{booktabs}
\usepackage{multirow}
\usepackage{microtype}
\usepackage{url}
\usepackage[hidelinks]{hyperref}
\usepackage[capitalize,noabbrev]{cleveref}
\usepackage{xcolor}
\usepackage{enumitem}
\usepackage{array}
\usepackage{algorithm}
\usepackage{algpseudocode}
\usepackage{tikz}
\usetikzlibrary{positioning,arrows.meta,fit,backgrounds}
\usepackage{pgfplots}
\pgfplotsset{compat=1.18}

\theoremstyle{plain}
\newtheorem{proposition}{Proposition}
\newtheorem{corollary}{Corollary}
\theoremstyle{definition}

\theoremstyle{remark}
\newtheorem{remark}{Remark}

\newcommand{\ToponeCovAlphaFive}{0.645}
\newcommand{\ToponeSizeAlphaFive}{0.68}
\newcommand{\ToponePrecAlphaFive}{0.948}
\newcommand{\ToponeRefCovAlphaFive}{0.286}
\newcommand{\ToponeDecAlphaFive}{0.825}
\newcommand{\FixedCovAlphaFive}{0.752}
\newcommand{\FixedSizeAlphaFive}{1.66}
\newcommand{\FixedPrecAlphaFive}{0.601}
\newcommand{\FixedRefCovAlphaFive}{0.542}
\newcommand{\FixedDecAlphaFive}{0.764}
\newcommand{\SuppoCovAlphaFive}{0.966}
\newcommand{\SuppoSizeAlphaFive}{4.87}
\newcommand{\SuppoPrecAlphaFive}{0.332}
\newcommand{\SuppoRefCovAlphaFive}{0.000}
\newcommand{\SuppoDecAlphaFive}{0.991}
\newcommand{\CesCovAlphaFive}{0.966}
\newcommand{\CesSizeAlphaFive}{4.87}
\newcommand{\CesPrecAlphaFive}{0.332}
\newcommand{\CesRefCovAlphaFive}{0.960}
\newcommand{\CesDecAlphaFive}{0.915}
\newcommand{\ToponeCovAlphaTen}{0.645}
\newcommand{\ToponeSizeAlphaTen}{0.68}
\newcommand{\ToponePrecAlphaTen}{0.948}
\newcommand{\ToponeRefCovAlphaTen}{0.286}
\newcommand{\ToponeDecAlphaTen}{0.825}
\newcommand{\FixedCovAlphaTen}{0.752}
\newcommand{\FixedSizeAlphaTen}{1.66}
\newcommand{\FixedPrecAlphaTen}{0.601}
\newcommand{\FixedRefCovAlphaTen}{0.542}
\newcommand{\FixedDecAlphaTen}{0.764}
\newcommand{\SuppoCovAlphaTen}{0.921}
\newcommand{\SuppoSizeAlphaTen}{3.26}
\newcommand{\SuppoPrecAlphaTen}{0.445}
\newcommand{\SuppoRefCovAlphaTen}{0.000}
\newcommand{\SuppoAbstAlphaTen}{0.057}
\newcommand{\SuppoDecAlphaTen}{0.953}
\newcommand{\CesCovAlphaTen}{0.921}
\newcommand{\CesSizeAlphaTen}{3.26}
\newcommand{\CesPrecAlphaTen}{0.445}
\newcommand{\CesRefCovAlphaTen}{0.929}
\newcommand{\CesAbstAlphaTen}{0.004}
\newcommand{\CesDecAlphaTen}{0.898}
\newcommand{\CesDecNAlphaTen}{804}
\newcommand{\ToponeCovAlphaTwenty}{0.645}
\newcommand{\ToponeSizeAlphaTwenty}{0.68}
\newcommand{\ToponePrecAlphaTwenty}{0.948}
\newcommand{\ToponeRefCovAlphaTwenty}{0.286}
\newcommand{\ToponeDecAlphaTwenty}{0.825}
\newcommand{\FixedCovAlphaTwenty}{0.752}
\newcommand{\FixedSizeAlphaTwenty}{1.66}
\newcommand{\FixedPrecAlphaTwenty}{0.601}
\newcommand{\FixedRefCovAlphaTwenty}{0.542}
\newcommand{\FixedDecAlphaTwenty}{0.764}
\newcommand{\SuppoCovAlphaTwenty}{0.830}
\newcommand{\SuppoSizeAlphaTwenty}{2.03}
\newcommand{\SuppoPrecAlphaTwenty}{0.566}
\newcommand{\SuppoRefCovAlphaTwenty}{0.000}
\newcommand{\SuppoDecAlphaTwenty}{0.859}
\newcommand{\CesCovAlphaTwenty}{0.830}
\newcommand{\CesSizeAlphaTwenty}{2.03}
\newcommand{\CesPrecAlphaTwenty}{0.566}
\newcommand{\CesRefCovAlphaTwenty}{0.854}
\newcommand{\CesDecAlphaTwenty}{0.838}
\newcommand{\Ncal}{2000}
\newcommand{\Ntest}{2000}
\newcommand{\Ndistract}{6}
\newcommand{\DecSuppOnly}{803}
\newcommand{\DecRefOnly}{1}
\newcommand{\DecNei}{0}
\newcommand{\DecMajority}{99.9}
\newcommand{\SuppCovCoords}{(0.05,0.9665) (0.1,0.9210) (0.15,0.8734) (0.2,0.8304) (0.3,0.7054)}
\newcommand{\RefCovCoords}{(0.05,0.9599) (0.1,0.9290) (0.15,0.9006) (0.2,0.8538) (0.3,0.7444)}
\newcommand{\CesFrontier}{(1.489,0.7054) (2.029,0.8304) (2.409,0.8734) (3.260,0.9210) (4.870,0.9665)}
\newcommand{\FixedSweepFrontier}{(1.282,0.6318) (1.506,0.7109) (1.655,0.7519) (1.778,0.7849) (1.913,0.8134) (2.054,0.8314) (2.229,0.8514) (2.499,0.8779) (2.722,0.8949) (3.071,0.9135) (3.857,0.9435)}
\newcommand{\ToponeSweepFrontier}{(0.641,0.6118) (0.662,0.6298) (0.679,0.6448) (0.693,0.6538) (0.704,0.6603)}
\newcommand{\AblateCovDZero}{0.902}
\newcommand{\AblateSizeDZero}{1.91}
\newcommand{\AblatePrecDZero}{0.603}
\newcommand{\AblateAbstDZero}{0.011}
\newcommand{\AblateCovDSix}{0.921}
\newcommand{\AblateSizeDSix}{3.26}
\newcommand{\AblatePrecDSix}{0.445}
\newcommand{\AblateAbstDSix}{0.004}
\newcommand{\BmCovDSix}{0.921}
\newcommand{\BmSizeDSix}{3.58}
\newcommand{\BmPrecDSix}{0.433}
\newcommand{\BmAbstDSix}{0.004}
\newcommand{\MsNSeeds}{5}
\newcommand{\MsDecMean}{0.874}
\newcommand{\MsDecStd}{0.0117}
\newcommand{\MsFCovMeanAlphaTen}{0.886}
\newcommand{\MsFCovStdAlphaTen}{0.007}
\newcommand{\MsFRefCovMeanAlphaTen}{0.836}
\newcommand{\MsFRefCovStdAlphaTen}{0.032}
\newcommand{\MsRCovMeanAlphaTen}{0.873}
\newcommand{\MsRCovStdAlphaTen}{0.010}
\newcommand{\MsMaxCovStd}{0.036}
\newcommand{\MsNCells}{18}
\newcommand{\MsVMaxCovStd}{0.020}
\newcommand{\FvAlphaFiveCov}{0.914}
\newcommand{\FvAlphaFiveRefCov}{0.902}
\newcommand{\FvAlphaTenCov}{0.833}
\newcommand{\FvAlphaTenRefCov}{0.815}
\newcommand{\FvAlphaTenDec}{0.802}
\newcommand{\FvAlphaTwentyCov}{0.747}
\newcommand{\FvAlphaTwentyRefCov}{0.673}
\newcommand{\FvDecN}{1500}
\newcommand{\FvDecSupp}{479}
\newcommand{\FvDecRef}{502}
\newcommand{\FvDecNei}{519}
\newcommand{\FvVerdToponeCov}{0.808}
\newcommand{\FvVerdToponeRefCov}{0.697}
\newcommand{\FvVerdToponeDec}{0.805}
\newcommand{\FvVerdToponeAbstPrec}{0.754}
\newcommand{\FvVerdFixedCov}{0.816}
\newcommand{\FvVerdFixedRefCov}{0.791}
\newcommand{\FvVerdFixedDec}{0.797}
\newcommand{\FvVerdFixedAbstPrec}{0.715}
\newcommand{\FvVerdSuppoCov}{0.833}
\newcommand{\FvVerdSuppoRefCov}{0.000}
\newcommand{\FvVerdSuppoDec}{0.577}
\newcommand{\FvVerdSuppoAbstPrec}{0.451}
\newcommand{\FvVerdCesCov}{0.833}
\newcommand{\FvVerdCesRefCov}{0.815}
\newcommand{\FvVerdCesDec}{0.802}
\newcommand{\FvVerdCesAbstPrec}{0.733}
\newcommand{\NliCovAlphaFive}{0.936}
\newcommand{\NliCovAlphaTen}{0.877}
\newcommand{\NliSizeAlphaTen}{6.71}
\newcommand{\NliDecAlphaTen}{0.516}
\newcommand{\NliCovAlphaTwenty}{0.769}
\newcommand{\FvStratMinCal}{30}
\newcommand{\FvStratGlobalCovFive}{0.914}
\newcommand{\FvStratCovFive}{0.946}
\newcommand{\FvStratDevFive}{0.4}
\newcommand{\FvStratGlobalCovTen}{0.833}
\newcommand{\FvStratCovTen}{0.869}
\newcommand{\FvStratDevTen}{3.1}
\newcommand{\FvStratGlobalCovTwenty}{0.747}
\newcommand{\FvStratCovTwenty}{0.785}
\newcommand{\FvStratDevTwenty}{1.5}
\newcommand{\UnivVcExistMin}{0.921}
\newcommand{\UnivVcExistMax}{0.995}
\newcommand{\UnivVcUnivMin}{0.457}
\newcommand{\UnivVcUnivMax}{0.916}
\newcommand{\UnivVcSizeMin}{3.26}
\newcommand{\UnivVcSizeMax}{8.31}
\newcommand{\UnivFvExistMin}{0.833}
\newcommand{\UnivFvExistMax}{0.935}
\newcommand{\UnivFvUnivMin}{0.766}
\newcommand{\UnivFvUnivMax}{0.879}
\newcommand{\UnivFvSizeMin}{0.64}
\newcommand{\UnivFvSizeMax}{1.14}
\newcommand{\DlDebSelDZero}{0.901}
\newcommand{\DlDebSelDFifty}{0.929}
\newcommand{\DlRobSelDZero}{0.516}
\newcommand{\DlRobSelDFifty}{0.363}
\newcommand{\CorrDeb}{+0.33}
\newcommand{\CorrRob}{-0.20}
\newcommand{\JointN}{1196}
\newcommand{\JointCovAlphaFive}{0.921}
\newcommand{\JointSuppAlphaFive}{0.960}
\newcommand{\JointRefAlphaFive}{0.960}
\newcommand{\JointCovAlphaTen}{0.843}
\newcommand{\JointCovAlphaTwenty}{0.694}
\newcommand{\XdCesCovVV}{0.919}
\newcommand{\XdCesCovVF}{0.935}
\newcommand{\XdCesCovFF}{0.833}
\newcommand{\XdCesCovFV}{0.779}
\newcommand{\GcPageMin}{0.830}
\newcommand{\GcPageMax}{0.963}
\newcommand{\GcPageNGroups}{13}
\newcommand{\FvAbstPrec}{0.733}
\newcommand{\FvAbstRec}{0.728}
\newcommand{\EpN}{194}
\newcommand{\EpMeta}{170}
\newcommand{\EpSuppElig}{18}
\newcommand{\EpSuppCov}{0.94}
\newcommand{\EpRefElig}{36}
\newcommand{\EpRefCov}{0.97}
\newcommand{\EpFalseSupp}{0.545}
\newcommand{\EpFree}{142}

\pgfplotsset{
  every axis/.append style={
    font=\scriptsize,
    label style={font=\scriptsize},
    tick label style={font=\scriptsize},
    legend style={font=\scriptsize, draw=none, fill=none, inner sep=1pt},
    legend cell align=left,
  },
}

\newcommand{\R}{\mathcal{R}}
\newcommand{\C}{\mathcal{C}}

\newcommand{\calib}{\mathcal{I}_{\mathrm{cal}}}
\newcommand{\supp}{\ensuremath{\mathrm{supp}}}
\newcommand{\refute}{\ensuremath{\mathrm{refute}}}
\newcommand{\abstain}{\ensuremath{\mathrm{abstain}}}

\usepackage{eso-pic}

\begin{document}

\title{Two-Sided Conformal Evidence Sets for Retrieval-Augmented Generation}

\author{\IEEEauthorblockN{Ben Jenkins}
\IEEEauthorblockA{\textit{Dept.\ of Electrical Engineering and Computer Science}\\
\textit{Florida Atlantic University}\\
Boca Raton, FL, USA\\
benjaminjenk2024@fau.edu}
\and
\IEEEauthorblockN{Mihaela Cardei}
\IEEEauthorblockA{\textit{Dept.\ of Electrical Engineering and Computer Science}\\
\textit{Florida Atlantic University}\\
Boca Raton, FL, USA\\
mcardei@fau.edu}}

\maketitle
\AddToShipoutPictureFG*{\AtPageLowerLeft{\put(\LenToUnit{0.5\paperwidth},\LenToUnit{0.32in}){\makebox(0,0)[b]{\parbox{\textwidth}{\scriptsize\centering \textcopyright~2026 IEEE. Personal use of this material is permitted. Permission from IEEE must be obtained for all other uses, in any current or future media, including reprinting/republishing this material for advertising or promotional purposes, creating new collective works, for resale or redistribution to servers or lists, or reuse of any copyrighted component of this work in other works.}}}}}

\begin{abstract}
Retrieval-augmented generation (RAG) can surface passages that support a claim, contradict it, or fail to address it, yet existing calibrated grounding methods are typically one-sided and model only supporting evidence. We introduce \emph{Conformal Evidence Sets} (CES), a two-sided framework that returns separate support and refutation sets with finite-sample inclusion coverage of at least $1-\alpha$ on each side under exchangeability; a Bonferroni union yields joint coverage of at least $1-2\alpha$. On labeled candidate pools from the contrastive VitaminC benchmark, CES closely tracks nominal coverage across $\alpha\!\in\!\{0.05,0.10,0.15,0.20,0.30\}$, with maximum deviations of $3.0$ percentage points for support and $5.4$ for refutation, and reaches essentially the same empirical coverage--size frontier as a test-tuned fixed-threshold heuristic without any test exposure. Stress tests mark the boundary of that guarantee: under FEVER distribution shift both sides under-cover, by up to $12.7$ percentage points, and support-side group-stratified recalibration substantially reduces the support gap. Scorer swaps, hard-negative retrieval, cross-dataset transfer, and a live end-to-end pilot expose a clear separation: when exchangeability holds and relevant evidence is retrieved, evidence inclusion remains calibrated even with an imperfect scorer, whereas the support/refute verdict tracks scorer quality directly. Calibrated evidence coverage and reliable verdicts are therefore distinct properties and should be reported separately.
\end{abstract}

\begin{IEEEkeywords}
retrieval-augmented generation, conformal prediction, uncertainty quantification, evidence attribution, hallucination detection, trustworthy AI
\end{IEEEkeywords}

\section{Introduction}
Retrieval-augmented generation (RAG) grounds language-model outputs in retrieved documents \cite{lewis2020rag,karpukhin2020dpr}, but retrieval alone does not make a claim trustworthy: retrieved passages may support the claim, contradict it, or fail to address it altogether. Recent work therefore separates retrieval and generation failures \cite{ru2024ragchecker} and distinguishes factual correctness from faithfulness to retrieved evidence \cite{fadeeva2025franq,wallat2025correctness}. Yet calibrated grounding methods are typically one-sided: they quantify whether supporting evidence is present without simultaneously representing contradictory evidence.

We introduce \emph{Conformal Evidence Sets} (CES), a claim-level grounding framework that calibrates both sides. The central contribution is a \emph{two-sided contrastive guarantee}: for each claim $c$, CES returns a support set $\widehat E^+_\alpha(c)$ and a refutation set $\widehat E^-_\alpha(c)$ satisfying
\[
\Pr[\widehat E^+_\alpha(c)\cap E^+(c)\neq\emptyset]\ge 1-\alpha
\]
and the analogous bound for refutation under exchangeability. A Bonferroni union yields a joint $1-2\alpha$ guarantee that both sides contain valid evidence. A method calibrating only support cannot provide both side-specific guarantees, which matters on contrastive examples where different passages support and refute the same claim.

Our target is inclusion of \emph{at least one} valid passage on each applicable side. This is well suited to human triage: when the retrieved pool contains sufficient evidence \cite{joren2025sufficient}, the reviewer wants at least one trustworthy item to inspect rather than a guarantee that every returned passage is correct. CES therefore returns structured support/refute/abstain evidence rather than treating unsupported truth as grounded truth.

A central distinction emerges from both the theory and the experiments: \emph{coverage is guaranteed; verdicts are earned}. Under exchangeability, and when relevant evidence is present in the candidate pool, evidence-set inclusion is controlled by conformal calibration for any scorer that produces support and contradiction scores, however weak. The final support/refute/abstain verdict, by contrast, depends on the quality of the underlying scorer and cannot be repaired downstream. Our scorer-swap, margin, FEVER, transfer, retrieval-noise, and end-to-end stress tests show that these two properties move independently. Coverage remains stable under scorer and distractor perturbations that leave the calibration/test sampling scheme unchanged, and degrades under transfer in a manner consistent with exchangeability violation; verdict quality, by contrast, follows scorer quality and distribution shift.

Our contributions are:
\begin{itemize}[leftmargin=*]
    \item A claim-level two-sided conformal formulation producing separate support and refutation evidence sets with finite-sample inclusion guarantees (\cref{sec:method,prop:coverage}).
    \item A support/refute/abstain decision rule that explicitly represents contrastive evidence while separating calibrated evidence inclusion from empirical verdict accuracy (\cref{sec:decision}).
    \item A comprehensive VitaminC evaluation showing near-nominal two-sided coverage, competitive coverage--size efficiency, robustness across scorers and retrieval noise, universal-coverage trade-offs, and stable multi-seed behavior (\cref{sec:experiments}).
    \item Stress tests on FEVER, cross-dataset transfer, and a live retrieve--generate--decompose pilot that identify exchangeability, scorer quality, and retrieval quality as distinct deployment failure modes (\cref{sec:fever,sec:transfer,sec:e2e}).
\end{itemize}

\section{Related Work}
RAG combines retrieval with generation for knowledge-intensive tasks \cite{lewis2020rag,karpukhin2020dpr,izacard2021fid}, creating a modular reliability problem spanning retrieval, context sufficiency, generation, and attribution. Recent evaluation work therefore separates retrieval and generation failures \cite{ru2024ragchecker}, studies whether retrieved context is sufficient \cite{joren2025sufficient}, and distinguishes factual correctness from faithfulness to the retrieved evidence \cite{fadeeva2025franq,wallat2025correctness}.

Conformal prediction provides distribution-free uncertainty sets under exchangeability \cite{vovk2005algorithmic,angelopoulos2023conformal}. In language modeling and RAG, prior work calibrates candidate generations, factuality back-off, response risk, or answer-level factuality \cite{quach2024conformal,mohri2024conformal,kang2024crag,feng2025conformalrag}. CES differs by calibrating \emph{claim-level, two-sided evidence sets}: one for support and one for refutation, each with its own inclusion guarantee.

Attributed-generation work such as ALCE and FActScore evaluate whether claims are supported by cited or retrieved evidence and exposes errors hidden by answer-level metrics \cite{gao2023alce,min2023factscore}. CES adopts this claim-level granularity but makes the surfaced evidence set itself the calibrated object.

\section{Problem Formulation}
Let $q$ be a user question and let $\R(q)=\{r_1,\ldots,r_k\}$ be the set of retrieved passages. A generator produces an answer $a$. A claim extractor decomposes $a$ into atomic claims $\C(a)=\{c_1,\ldots,c_m\}$. For each claim $c_j$, the goal is to output a pair of evidence sets
\[
\widehat{E}^{+}_{\alpha}(c_j),\ \widehat{E}^{-}_{\alpha}(c_j) \subseteq \R(q)
\]
and a decision
\[
y_j \in \{\supp, \refute, \abstain\}.
\]
Each evidence set should be small enough for human inspection but large enough to include at least one truly supporting (respectively refuting) passage when such a passage exists in the retrieved pool.

We distinguish three grounding outcomes. A claim is \emph{faithfully supported} if the retrieved passages contain enough evidence to entail the claim. A claim is \emph{refuted} if retrieved passages contain enough evidence to contradict it. A claim is \emph{ungrounded} if it may be factually true in the world but is not supported by the retrieved context; the framework treats these as distinct outcomes throughout.

\section{Method}\label{sec:method}

\begin{figure}[t]
\centering
\begin{tikzpicture}[
  node distance=4pt and 6pt,
  every node/.style={font=\scriptsize},
  frontend/.style={draw, rounded corners=2pt, align=center,
                   minimum height=18pt, inner sep=3pt,
                   fill=gray!8, draw=gray!50, dashed,
                   text=gray!50!black},
  block/.style={draw, rounded corners=2pt, align=center, line width=0.8pt,
                minimum height=18pt, inner sep=3pt,
                fill=blue!10, draw=blue!60},
  data/.style={draw, align=center, minimum height=18pt, inner sep=3pt,
               fill=orange!10, draw=orange!60, line width=0.8pt},
  arr/.style={->, >=Stealth, thick, shorten >=1pt, shorten <=1pt},
  arrf/.style={->, >=Stealth, thick, shorten >=1pt, shorten <=1pt,
               gray!50!black, dashed},
]
\node[frontend] (q) {Question $q$};
\node[frontend, right=of q] (ret) {Retrieve\\ top-$k$};
\node[frontend, right=of ret] (gen) {Generate};
\node[frontend, below=14pt of gen] (dec) {Decompose\\ into claims};
\node[block, left=of dec] (nli) {\textbf{NLI score}\\ (entail, contra)};
\node[data, left=of nli] (cal) {\textbf{Calibrate}\\ $\tau^+_\alpha$, $\tau^-_\alpha$};
\node[data, below=14pt of nli, fill=green!12, draw=green!60, line width=0.8pt] (out) {\textbf{Per-claim verdict:}\\ supp~/~refute~/~abstain};
\draw[arrf] (q) -- (ret);
\draw[arrf] (ret) -- (gen);
\draw[arrf] (gen) -- (dec);
\draw[arrf] (dec) -- (nli);
\draw[arrf] (ret) to[bend right=22] (nli);
\draw[arr] (cal) -- (nli);
\draw[arr] (nli) -- (out);
\end{tikzpicture}
\caption{Conformal Evidence Sets pipeline. The calibration-and-decision layer (NLI scoring, $\tau^{\pm}_\alpha$, verdict) is what our main experiments evaluate (\Cref{sec:scope}); the retrieve--generate--decompose front end (dashed, grey) is exercised only by the end-to-end pilot of \cref{sec:e2e}.}
\label{fig:pipeline}
\end{figure}
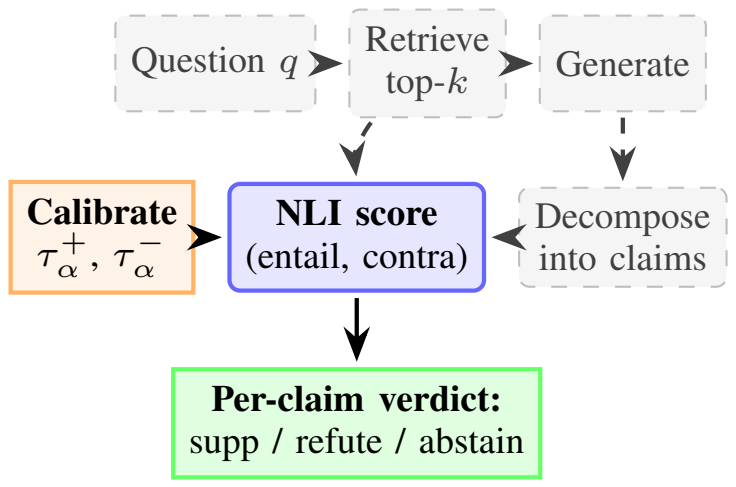

\Cref{fig:pipeline} summarizes how the framework fits into a RAG stack. The retrieve--generate--decompose front end is standard and can be swapped freely; our contribution is the calibration-and-decision layer, i.e.\ split-conformal calibration and the support/refute/abstain decision rule operating over standard NLI scores, which is what the experiments evaluate.

\subsection{Claim Decomposition}
We assume each input is a single atomic claim $c$ paired with a candidate evidence pool $\R(q)$. When the input is a longer generated answer, an upstream claim extractor (an LLM with a constrained prompt, or a supervised claim-segmentation model) splits the answer into atomic claims first. The experiments in \Cref{sec:experiments} use VitaminC, whose claims are already atomic, so this step is not exercised.

\subsection{Evidence Scoring}
For each claim--passage pair $(c, r_i)$, an NLI scorer returns support and contradiction probabilities $p^+(c, r_i)$ and $p^-(c, r_i)$. The nonconformity scores are
\[
A^+(c, r_i) = 1 - p^+(c, r_i),\qquad A^-(c, r_i) = 1 - p^-(c, r_i),
\]
so that low nonconformity corresponds to strong evidence. Other entailment scorers (cross-encoders, LLM judges) plug in without changing the conformal layer.

\subsection{Split-Conformal Evidence Sets}
Let the calibration set $\calib$ contain claims with annotated supporting and refuting passages, written $E^+(c_j)$ and $E^-(c_j)$, and let $\calib^{\pm} = \{j \in \calib : E^{\pm}(c_j) \neq \emptyset\}$ be the side-eligible subsets, with $n_{\pm} = |\calib^{\pm}|$. The per-claim calibration scores are
\begin{align*}
A^+_j &= \min_{r_i \in E^+(c_j)} A^+(c_j, r_i),\quad j \in \calib^{+};\\
A^-_j &= \min_{r_i \in E^-(c_j)} A^-(c_j, r_i),\quad j \in \calib^{-},
\end{align*}
so $A^+_j$ is the nonconformity of the most-entailing gold supporter (and analogously for refutation). Let $k_{\pm} = \lceil(n_{\pm}+1)(1-\alpha)\rceil$ and let $A^{\pm}_{(1)} \le \cdots \le A^{\pm}_{(n_{\pm})}$ denote the ordered calibration scores. The split-conformal thresholds are the order statistics
\[
\tau^{\pm}_{\alpha} =
\begin{cases}
A^{\pm}_{(k_{\pm})}, & k_{\pm} \le n_{\pm},\\
\infty, & k_{\pm} = n_{\pm}+1.
\end{cases}
\]
For a new claim $c$ with retrieved pool $\R(q)$,
\[
\widehat{E}^{\pm}_{\alpha}(c) = \{r_i \in \R(q) : A^{\pm}(c, r_i) \leq \tau^{\pm}_\alpha\}.
\]
Each side is calibrated separately because the supporting and refuting nonconformity scores follow different distributions; the same calibration split can be used for both sides, with each threshold computed from its side-eligible subset. \Cref{alg:ces} gives the complete procedure.

\begin{algorithm}[t]
\caption{Conformal Evidence Sets}
\label{alg:ces}
\begin{algorithmic}[1]
\Require Calibration claims $\{(c_j, E^+(c_j), E^-(c_j))\}_{j\in\calib}$ with annotated supporting set $E^+(c_j)$ and refuting set $E^-(c_j)$; test claim $c$ with retrieved pool $\R(q)$; NLI scorer giving $p^+(c, r), p^-(c, r)$; level $\alpha$; margin $\delta$.
\State $A^+_j \gets \min_{r \in E^+(c_j)}\,(1 - p^+(c_j, r))$ for each $j \in \calib^{+}$
\State $A^-_j \gets \min_{r \in E^-(c_j)}\,(1 - p^-(c_j, r))$ for each $j \in \calib^{-}$
\State $\tau^+_\alpha \gets A^+_{(k_+)}$, the $k_+$-th order statistic with $k_+ = \lceil(n_+ + 1)(1-\alpha)\rceil$ ($\infty$ if $k_+ > n_+$)
\State $\tau^-_\alpha \gets A^-_{(k_-)}$, analogously
\State for each $r_i \in \R(q)$: $a^+_i \gets 1 - p^+(c, r_i)$, $a^-_i \gets 1 - p^-(c, r_i)$
\State $\widehat{E}^+ \gets \{r_i : a^+_i \leq \tau^+_\alpha\}$
\State $\widehat{E}^- \gets \{r_i : a^-_i \leq \tau^-_\alpha\}$
\If{$\widehat{E}^+ = \emptyset$ and $\widehat{E}^- = \emptyset$} \Return $(\abstain, \widehat{E}^+, \widehat{E}^-)$ \Comment{neither side calibrated; nothing to dominate}
\ElsIf{$\widehat{E}^+ \neq \emptyset$ and $\widehat{E}^- = \emptyset$} \Return $(\supp, \widehat{E}^+, \widehat{E}^-)$
\ElsIf{$\widehat{E}^- \neq \emptyset$ and $\widehat{E}^+ = \emptyset$} \Return $(\refute, \widehat{E}^+, \widehat{E}^-)$
\EndIf
\State $\Delta \gets \max_{r_i \in \widehat{E}^+}(1 - a^+_i) - \max_{r_i \in \widehat{E}^-}(1 - a^-_i)$ \Comment{both non-empty; dominance margin well-defined}
\If{$\Delta > \delta$} \Return $(\supp, \widehat{E}^+, \widehat{E}^-)$
\ElsIf{$-\Delta > \delta$} \Return $(\refute, \widehat{E}^+, \widehat{E}^-)$
\Else{} \Return $(\abstain, \widehat{E}^+, \widehat{E}^-)$
\EndIf
\end{algorithmic}
\end{algorithm}

\subsection{Decision Rule}\label{sec:decision}
The verdict is defined procedurally to avoid undefined arithmetic on empty sets. (i) If both $\widehat{E}^+$ and $\widehat{E}^-$ are empty, CES abstains. (ii) If exactly one is non-empty, CES returns that side. (iii) If both are non-empty, CES computes the dominance margin
\[
\Delta(c) = \max_{r_i \in \widehat{E}^+_{\alpha}(c)} \!\!(1 - A^+(c, r_i))\;-\;
           \max_{r_i \in \widehat{E}^-_{\alpha}(c)} \!\!(1 - A^-(c, r_i))
\]
and returns \supp{} when $\Delta(c) > \delta$, \refute{} when $-\Delta(c) > \delta$, and \abstain{} otherwise. The hyperparameter $\delta \geq 0$ controls how confidently one side must dominate; experiments use $\delta = 0$, and a sweep in \cref{sec:nliswap} shows it is the best default among the settings tested ($\delta$ acts as a selective-prediction knob only when the underlying scorer is strong). The verdict is post-hoc with respect to the calibrated thresholds, so its accuracy is not a conformal quantity and is measured empirically (\Cref{rem:verdict}).

\subsection{Coverage Guarantee}
\begin{proposition}\label{prop:coverage}
Fix $\alpha \in (0, 1)$ and an evidence-scoring function chosen independently of the calibration and test data. Suppose the side-eligible calibration tuples of (claim, retrieved pool, gold evidence) and the test tuple are exchangeable after applying the same eligibility rule, i.e.\ the test claim, like every claim in $\calib^{+}$, has at least one annotated supporting passage in its retrieved pool. Let $\tau^+_\alpha = A^+_{(k_+)}$ be the order-statistic threshold above with $k_{+} = \lceil(n_{+}+1)(1-\alpha)\rceil$ (and $\tau^+_\alpha = \infty$ when $k_{+} > n_{+}$, where the guarantee holds trivially), and let $\widehat{E}^+_\alpha(c) = \{r_i \in \R(q) : A^+(c, r_i) \leq \tau^+_\alpha\}$. Then
\[
\Pr\bigl(\widehat{E}^+_\alpha(c) \cap E^+(c) \neq \emptyset\bigr) \geq 1 - \alpha,
\]
where $E^+(c)$ is the set of true supporting passages for $c$. The analogous bound holds for the refutation set $\widehat{E}^-_\alpha(c)$ under the parallel exchangeability assumption on refuting passages.
\end{proposition}

\begin{proof}[Proof sketch]
For each calibration claim $j$, let $A^+_j = \min_{r_i \in E^+(c_j)} A^+(c_j, r_i)$, and define $A^+_\star = \min_{r_i \in E^+(c)} A^+(c, r_i)$ for the test claim. By exchangeability of the claim--pool--evidence tuples and the fact that the scoring function is fixed independently of them, $\{A^+_j\}_{j\in\calib^{+}} \cup \{A^+_\star\}$ is an exchangeable random sequence. Standard split-conformal arguments \cite{vovk2005algorithmic,angelopoulos2023conformal} give
\[
\Pr(A^+_\star \leq \tau^+_\alpha) \geq 1 - \alpha.
\]
The event $A^+_\star \leq \tau^+_\alpha$ implies that at least one $r_i \in E^+(c)$ has $A^+(c, r_i) \leq \tau^+_\alpha$, hence lies in $\widehat{E}^+_\alpha(c)$. The refutation case is identical by symmetry.
\end{proof}

\begin{corollary}[Joint two-sided coverage]\label{cor:joint}
If the support and refutation calibration assumptions both hold (in particular, the test claim carries gold evidence on both sides), then by the union bound,
\[
\Pr\!\left(
\widehat{E}^{+}_{\alpha}(c)\cap E^{+}(c)\neq\emptyset
\ \text{and}\
\widehat{E}^{-}_{\alpha}(c)\cap E^{-}(c)\neq\emptyset
\right)
\geq 1-2\alpha.
\]
To target joint coverage at level $1-\alpha$, calibrate each side at level $\alpha/2$.
\end{corollary}

\begin{proposition}[Universal coverage variant]\label{prop:universal}
\Cref{prop:coverage} guarantees inclusion of \emph{at least one} gold passage. For applications that consume the full evidence set (e.g., citation rendering, audit), a stricter guarantee is available: replace $A^+_j = \min_{r_i \in E^+(c_j)} A^+(c_j, r_i)$ in the proof by $A^+_j = \max_{r_i \in E^+(c_j)} A^+(c_j, r_i)$. The resulting threshold $\tau^{+,\mathrm{uni}}_\alpha$ satisfies
\[
\Pr\bigl(\widehat{E}^{+,\mathrm{uni}}_\alpha(c) \supseteq E^+(c)\bigr) \geq 1 - \alpha
\]
by an identical split-conformal argument. The two variants coincide when $|E^+(c)| = 1$; for multi-evidence claims the universal variant gives a weakly larger (and typically strictly larger) set. The choice is a calibration-time knob, not a re-implementation. \Cref{tab:univcov} reports the empirical trade-off.
\end{proposition}

\begin{remark}[Coverage versus verdict accuracy]\label{rem:verdict}
The guarantee applies to evidence-set inclusion, not to the downstream support/refute/abstain verdict. The verdict additionally depends on the dominance margin between support and refutation strengths in \Cref{sec:decision}, so a claim with non-empty $\widehat{E}^+$ can still be reported as \refute{} or \abstain{} if the refutation side dominates. We therefore report set coverage and verdict accuracy separately in the experiments.
\end{remark}

\begin{remark}[Inclusion is not truth]\label{rem:not-truth}
\Cref{prop:coverage} guarantees \emph{inclusion of available support}, not \emph{truth in the world}. If retrieval omits every truly supporting passage, the premise of the coverage statement is not satisfied and no calibration procedure can recover the missing evidence. CES may then abstain, but it may also return a support or refutation verdict when retrieved non-evidence receives sufficiently strong scores (the end-to-end pilot of \cref{sec:e2e} exhibits exactly this behavior). The guarantee is therefore conditional on relevant evidence being present in the candidate pool and must not be read as a factual-correctness guarantee.
\end{remark}

\paragraph{Worked example.} A VitaminC claim states that fewer than $40{,}626$ coronavirus cases had been confirmed on 10 February 2020. Two revisions of the same paragraph report $40{,}625$ and $40{,}627$ cases, respectively. At $\alpha=0.10$, CES places the former in $\widehat E^+$ and the latter in $\widehat E^-$, illustrating the contrastive evidence that a one-sided method must suppress.

\subsection{Scope of Evaluation}\label{sec:scope}
Main experiments isolate the calibration-and-decision layer on VitaminC's labeled candidate pools, bypassing retrieval, generation, and decomposition so measured differences are attributable to calibration rather than a particular front end. \Cref{sec:e2e} adds a small live retrieve--generate--decompose pilot as a scope check; full end-to-end evaluation remains future work.

\section{Experiments}\label{sec:experiments}

\subsection{Setup}\label{sec:setup}
\textbf{Benchmark.} We evaluate on VitaminC \cite{schuster2021vitaminc}, the natural benchmark for two-sided evidence sets: the \emph{same} claim is paired with both supporting and refuting evidence drawn from different revisions of the same Wikipedia paragraph, so each \texttt{case\_id} yields a candidate pool mixing \textsc{supports}, \textsc{refutes}, and \textsc{not enough info} passages on one topic, and a one-sided calibrated method will systematically miss a side. Other fact-verification benchmarks (e.g.\ FEVER \cite{thorne2018fever}) typically pair each claim with one-sided evidence and cannot expose this failure mode.

\textbf{Pools.} For each grouped claim we use the labeled VitaminC passages as the in-domain pool and append $\Ndistract$ distractor passages sampled from other case ids as generic retrieval noise (\cref{sec:distract} additionally evaluates BM25-mined distractors that mimic lexical-retriever confusions). We calibrate on $\Ncal$ claims from the training split and evaluate on $\Ntest$ disjoint claims from the validation split.

\textbf{NLI scorer.} Support and contradiction probabilities $p^+(c, r_i), p^-(c, r_i)$ are produced by a DeBERTa-v3 NLI model fine-tuned on a mixture that includes MNLI, ANLI, and FEVER \cite{he2021debertav3,williams2018mnli,nie2020anli,thorne2018fever}. Nonconformity scores are $A^+ = 1 - p^+$, $A^- = 1 - p^-$.

\textbf{Baselines.} We compare four claim-level passage-selection methods that share the same NLI scores so the comparison isolates the calibration strategy:
\begin{itemize}[leftmargin=*]
    \item \textbf{Top-1 NLI:} let $r^+$ ($r^-$) be the passage maximizing $p^+$ ($p^-$); abstain if $\max(p^+(r^+), p^-(r^-)) < 0.5$, otherwise return the stronger side with that single passage.
    \item \textbf{Fixed threshold:} include every passage with $p^+ \geq 0.7$ in the support set and every passage with $p^- \geq 0.7$ in the refutation set; return the side with more passages (ties favor \supp{}), abstaining only when both sets are empty.
    \item \textbf{Conformal (support only):} a single-sided conformal evidence set with calibrated $\tau^+_\alpha$; abstain only when $\widehat{E}^+ = \emptyset$. Mirrors single-sided conformal RAG methods.
    \item \textbf{Conformal Evidence Sets (ours):} two-sided sets $\widehat{E}^+, \widehat{E}^-$ with the support/refute/abstain decision rule of \Cref{sec:decision}.
\end{itemize}

\textbf{Metrics.}
\emph{Support coverage} is the fraction of claims whose gold supporting passages intersect $\widehat{E}^+$, restricted to claims that have gold support; the analogous quantity is reported for refutation. \emph{Support size} is the mean cardinality of $\widehat{E}^+$. \emph{Support precision} is the mean fraction of $\widehat{E}^+$ that is gold-supporting, averaged over claims with gold support and a non-empty predicted set (empty sets are skipped rather than scored). \emph{Frac.\ abstain} is the share of test claims for which the decision rule returns \abstain. \emph{Decision accuracy} is verdict-vs-gold agreement on claims with a unique gold label; contrastive claims, which carry gold evidence on both sides, are excluded because no single verdict is uniquely correct for them, leaving $\CesDecNAlphaTen$ evaluable claims in the main setup. This exclusion leaves a near-degenerate remainder on VitaminC: $\DecSuppOnly$ of those claims are gold-\supp{}, $\DecRefOnly$ is gold-\refute{}, and $\DecNei$ are gold-NEI, so a constant \supp{} predictor scores $\DecMajority\%$. We therefore report the column for completeness rather than as a comparison: on VitaminC the verdict task is informative only on the contrastive claims that the metric must exclude, and a one-sided method scores well precisely by never emitting \refute{}. \Cref{tab:fever-verdict} evaluates verdicts on FEVER, whose label mix is balanced.

\subsection{Coverage Tracks Nominal}\label{sec:coverage}
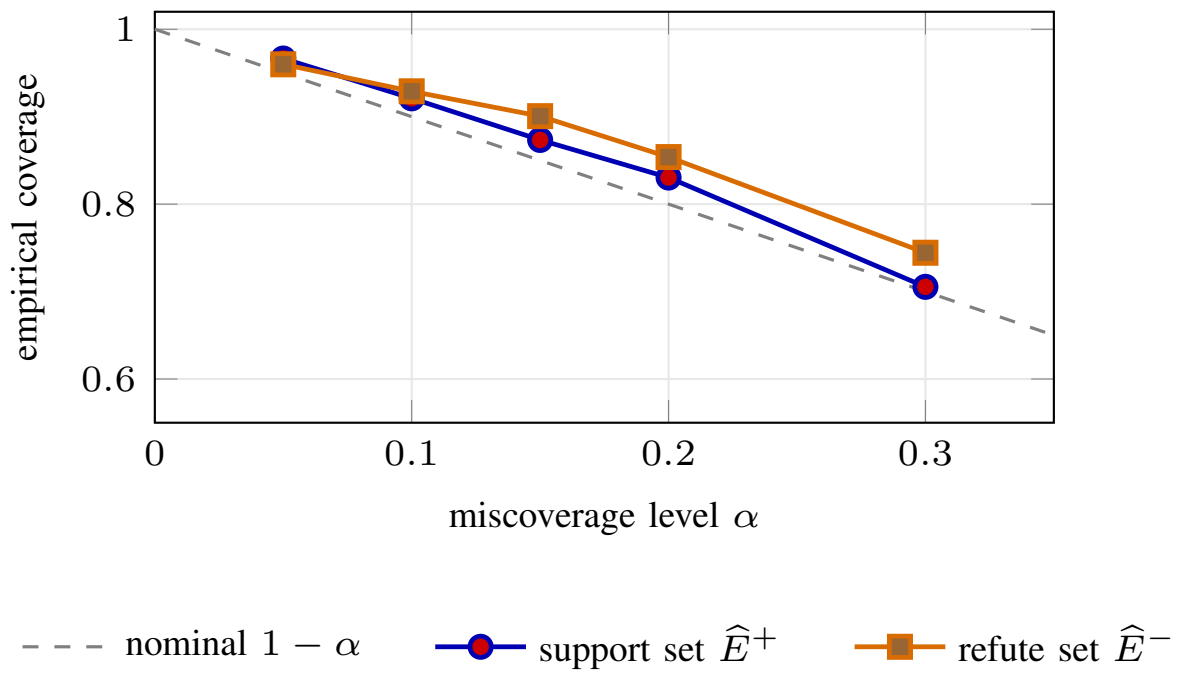
\begin{figure}[t]
\centering
\begin{tikzpicture}
\begin{axis}[
  width=0.85\linewidth, height=4.3cm,
  xlabel={miscoverage level $\alpha$},
  ylabel={empirical coverage},
  xmin=0, xmax=0.35, ymin=0.55, ymax=1.02,
  grid=both, grid style={gray!18},
  legend style={at={(0.5,-0.45)}, anchor=north, legend columns=-1,
                /tikz/every even column/.append style={column sep=0.4cm}},
  domain=0:0.35, samples=2,
]
\addplot+[dashed, gray, mark=none, line width=0.6pt] {1 - x};
\addlegendentry{nominal $1-\alpha$}
\addplot+[mark=*, blue!70!black, line width=0.9pt] coordinates {\SuppCovCoords};
\addlegendentry{support set $\widehat{E}^+$}
\addplot+[mark=square*, orange!85!black, line width=0.9pt] coordinates {\RefCovCoords};
\addlegendentry{refute set $\widehat{E}^-$}
\end{axis}
\end{tikzpicture}
\caption{Empirical coverage of conformal support and refutation evidence sets versus nominal $1-\alpha$ on VitaminC ($n_{\mathrm{cal}}=\Ncal$, $n_{\mathrm{test}}=\Ntest$). Both curves stay at or above the diagonal across all $\alpha$.}
\label{fig:coverage}
\end{figure}

\Cref{fig:coverage} plots empirical support and refutation coverage of Conformal Evidence Sets against the nominal target $1-\alpha$ over $\alpha \in \{0.05, 0.10, 0.15, 0.20, 0.30\}$. Support coverage stays close to nominal across the sweep (deviations $\{1.6, 2.1, 2.3, 3.0, 0.5\}$ percentage points), while refutation coverage is more conservative, with the largest deviation $\approx\!5.4$ points at $\alpha=0.20$. \Cref{prop:coverage} gives a marginal finite-sample guarantee of coverage at least $1-\alpha$ under exchangeability; the observed overshoot is unlikely to be explained by the finite-sample ceiling correction alone (which contributes at most $1/(n_{\mathrm{cal}}+1)\!\approx\!0.05$ percentage points at $n_{\mathrm{cal}}=\Ncal$). More plausible causes are discreteness and saturation in the NLI probabilities, especially near high entailment or contradiction scores, and heterogeneity in the number and difficulty of valid evidence passages per claim.

\paragraph{How conservative is the joint bound?} The Bonferroni union in \Cref{cor:joint} is conservative by construction, so we measure the joint contrastive event directly on the $\JointN$ test claims with gold evidence on both sides. The empirical joint rate exceeds the $1-2\alpha$ bound at every level ($\JointCovAlphaFive$ vs.\ $0.90$ at $\alpha=0.05$; $\JointCovAlphaTen$ vs.\ $0.80$; $\JointCovAlphaTwenty$ vs.\ $0.60$) and, at every level, matches the product of the two per-side rates on the same subset to within half a point (at $\alpha=0.05$: $\JointSuppAlphaFive \times \JointRefAlphaFive$ vs.\ observed $\JointCovAlphaFive$). The support and refutation inclusion events are thus empirically close to independence on this benchmark: the union bound is only mildly loose at practical operating points ($\alpha \le 0.10$), so a tighter joint scheme would buy little there.

\subsection{Calibrated vs.\ Uncalibrated Frontiers}
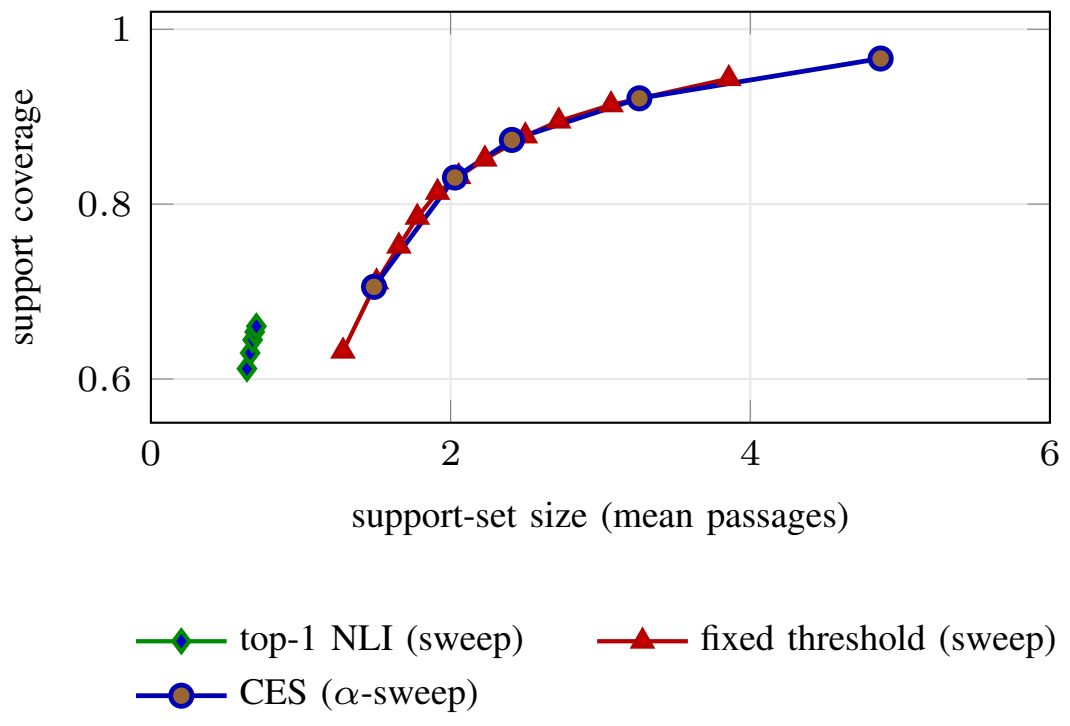
\begin{figure}[t]
\centering
\begin{tikzpicture}
\begin{axis}[
  width=0.85\linewidth, height=4.3cm,
  xlabel={support-set size (mean passages)},
  ylabel={support coverage},
  xmin=0, xmax=6, ymin=0.55, ymax=1.02,
  grid=both, grid style={gray!18},
  legend style={at={(0.5,-0.45)}, anchor=north, legend columns=2,
                /tikz/every even column/.append style={column sep=0.4cm}},
]
\addplot+[mark=diamond*, green!55!black, line width=0.8pt] coordinates {\ToponeSweepFrontier};
\addlegendentry{top-1 NLI (sweep)}
\addplot+[mark=triangle*, red!70!black, line width=0.8pt] coordinates {\FixedSweepFrontier};
\addlegendentry{fixed threshold (sweep)}
\addplot+[mark=*, blue!70!black, line width=0.9pt] coordinates {\CesFrontier};
\addlegendentry{CES ($\alpha$-sweep)}
\end{axis}
\end{tikzpicture}
\caption{Coverage--size frontiers on VitaminC after hyperparameter sweeps. Fixed-threshold (red) and CES (blue) overlap closely; the conformal frontier is reached without test exposure and carries a pointwise finite-sample guarantee at each prespecified $\alpha$.}
\label{fig:frontier}
\end{figure}

\Cref{fig:frontier} compares three of the four baselines after sweeping their tunable hyperparameters; the support-only conformal variant traces the same $\alpha$-sweep curve as CES on this axis (one-sided CES is, by construction, two-sided CES projected to the support side) and is omitted to avoid superposition. Fixed-threshold (entail probability cut-off $\theta \in \{0.05, \ldots, 0.90\}$) and top-1 NLI (abstain threshold $\theta \in \{0.3, \ldots, 0.7\}$) trace empirical frontiers; CES traces a frontier by sweeping $\alpha$. Top-1 sits in a tight precise corner ($\ToponePrecAlphaFive$ precision, $\ToponeCovAlphaFive$ coverage at most). Fixed-threshold and CES achieve essentially the same coverage--size frontier: at coverage $\approx 0.91$, fixed-threshold at $\theta=0.10$ uses on average $3.07$ passages while CES at $\alpha=0.10$ uses $\CesSizeAlphaTen$ passages.

The numerical equivalence is unsurprising, since split-conformal calibration is by construction the data-adaptive threshold whose calibration-set coverage equals $1-\alpha$; the distinction is what the curves \emph{cost} to obtain. The fixed-threshold frontier is an \emph{oracle diagnostic}, constructed by sweeping $\theta$ directly against evaluation data with known coverage, and is not a deployable baseline; CES's frontier requires a single calibration draw and inherits the distribution-free guarantee of \Cref{prop:coverage}. That even this oracle sweep yields no meaningfully better frontier is the point. Top-1 cannot reach the upper coverage regime at any threshold because it returns at most one passage. We therefore frame CES not as dominating a frontier, but as reaching the same frontier (i) without test exposure, (ii) with a pointwise finite-sample guarantee at each prespecified $\alpha$, and (iii) with the joint two-sided bound of \Cref{cor:joint} that no fixed-threshold rule provides.

\subsection{Headline Numbers}
\Cref{tab:main} reports the main metrics at three $\alpha$ levels. The two conformal variants share their support metrics by design (same $\tau^+_\alpha$); they diverge in the decision distribution, where two-sided evidence sets engage contrastive cases rather than defaulting to a one-sided verdict (abstention rate $\CesAbstAlphaTen$ for CES vs.\ $\SuppoAbstAlphaTen$ for support-only at $\alpha=0.10$).

\begin{table}[t]
\centering
\caption{Main VitaminC results ($n_{\mathrm{cal}}=\Ncal$, $n_{\mathrm{test}}=\Ntest$). Coverage targets are $1-\alpha$. Both conformal methods share the support-side guarantee; CES additionally calibrates refutation and supplies the joint two-sided bound. Dec.\ Acc.\ excludes contrastive claims, leaving a subset whose majority-class baseline is $\DecMajority\%$ (\cref{sec:setup}); \cref{tab:fever-verdict} compares verdicts where the label mix is balanced. CES results are replicated over $\MsNSeeds$ seeds (see Reproducibility).}
\label{tab:main}
\setlength{\tabcolsep}{2pt}
\begin{tabular}{llccccc}
\toprule
$\alpha$ & Method & Supp.\ Cov. & Ref.\ Cov. & Size & Prec. & Dec.\ Acc. \\
\midrule
\multirow{4}{*}{$0.05$}
  & Top-1 NLI       & \ToponeCovAlphaFive   & \ToponeRefCovAlphaFive & \ToponeSizeAlphaFive   & \ToponePrecAlphaFive   & \ToponeDecAlphaFive \\
  & Fixed thr.      & \FixedCovAlphaFive    & \FixedRefCovAlphaFive  & \FixedSizeAlphaFive    & \FixedPrecAlphaFive    & \FixedDecAlphaFive \\
  & Conf.\ supp-only& \SuppoCovAlphaFive    & \SuppoRefCovAlphaFive  & \SuppoSizeAlphaFive    & \SuppoPrecAlphaFive    & \SuppoDecAlphaFive \\
  & \textbf{CES (ours)} & \textbf{\CesCovAlphaFive} & \textbf{\CesRefCovAlphaFive} & \CesSizeAlphaFive & \CesPrecAlphaFive & \CesDecAlphaFive \\
\midrule
\multirow{4}{*}{$0.10$}
  & Top-1 NLI       & \ToponeCovAlphaTen    & \ToponeRefCovAlphaTen & \ToponeSizeAlphaTen   & \ToponePrecAlphaTen   & \ToponeDecAlphaTen \\
  & Fixed thr.      & \FixedCovAlphaTen     & \FixedRefCovAlphaTen  & \FixedSizeAlphaTen    & \FixedPrecAlphaTen    & \FixedDecAlphaTen \\
  & Conf.\ supp-only& \SuppoCovAlphaTen     & \SuppoRefCovAlphaTen  & \SuppoSizeAlphaTen    & \SuppoPrecAlphaTen    & \SuppoDecAlphaTen \\
  & \textbf{CES (ours)} & \textbf{\CesCovAlphaTen} & \textbf{\CesRefCovAlphaTen} & \CesSizeAlphaTen & \CesPrecAlphaTen & $\MsDecMean\!\pm\!\MsDecStd$ \\
\midrule
\multirow{4}{*}{$0.20$}
  & Top-1 NLI       & \ToponeCovAlphaTwenty & \ToponeRefCovAlphaTwenty & \ToponeSizeAlphaTwenty & \ToponePrecAlphaTwenty & \ToponeDecAlphaTwenty \\
  & Fixed thr.      & \FixedCovAlphaTwenty  & \FixedRefCovAlphaTwenty  & \FixedSizeAlphaTwenty  & \FixedPrecAlphaTwenty  & \FixedDecAlphaTwenty \\
  & Conf.\ supp-only& \SuppoCovAlphaTwenty  & \SuppoRefCovAlphaTwenty  & \SuppoSizeAlphaTwenty  & \SuppoPrecAlphaTwenty  & \SuppoDecAlphaTwenty \\
  & \textbf{CES (ours)} & \textbf{\CesCovAlphaTwenty} & \textbf{\CesRefCovAlphaTwenty} & \CesSizeAlphaTwenty & \CesPrecAlphaTwenty & \CesDecAlphaTwenty \\
\bottomrule
\end{tabular}
\end{table}

\subsection{Coverage Variant: Universal Inclusion}\label{sec:univcov}
The universal variant of \Cref{prop:universal} trades larger sets for inclusion of every gold passage. At $\alpha=0.10$ on VitaminC, it raises universal coverage from \UnivVcUnivMin{} to \UnivVcUnivMax{} while increasing mean support-set size from \UnivVcSizeMin{} to \UnivVcSizeMax{} passages ($2.5\times$); on FEVER, where multi-evidence claims are rarer, the coverage and size changes are much smaller (\UnivFvUnivMin{}$\to$\UnivFvUnivMax{}, \UnivFvSizeMin{}$\to$\UnivFvSizeMax{}). Thus existential calibration suits any-one-passage triage, whereas universal calibration suits audit or citation rendering.

\begin{table}[t]
\centering
\caption{Coverage variants at $\alpha=0.10$. Each row is a calibration variant ($\min$- vs.\ $\max$-reduction over the gold set); each is evaluated under both coverage notions, existential (at least one gold passage included) and universal (all included). Existential calibration cannot deliver universal coverage; universal calibration buys it with larger sets.}
\label{tab:univcov}
\setlength{\tabcolsep}{4pt}
\footnotesize
\begin{tabular}{lcccc}
\toprule
Dataset & Variant & Exist. cov. & Univ. cov. & Size \\
\midrule
\multirow{2}{*}{VitaminC}
 & Existential & \UnivVcExistMin & \UnivVcUnivMin & \UnivVcSizeMin \\
 & Universal   & \UnivVcExistMax & \UnivVcUnivMax & \UnivVcSizeMax \\
\midrule
\multirow{2}{*}{FEVER}
 & Existential & \UnivFvExistMin & \UnivFvUnivMin & \UnivFvSizeMin \\
 & Universal   & \UnivFvExistMax & \UnivFvUnivMax & \UnivFvSizeMax \\
\bottomrule
\end{tabular}
\end{table}

\subsection{Exchangeability Stress Test: FEVER Transfer}\label{sec:fever}
The coverage guarantee is conditional on calibration/test exchangeability. We therefore transfer CES to \texttt{copenlu/fever\_gold\_evidence} with $1500$ disjoint calibration and test claims and the same distractor augmentation. Unstratified support coverage falls $3$--$7$ pp below nominal across the tested levels (\FvAlphaFiveCov, \FvAlphaTenCov, \FvAlphaTwentyCov), and refutation under-covers even more severely (\FvAlphaFiveRefCov, \FvAlphaTenRefCov, \FvAlphaTwentyRefCov; up to $12.7$ pp at $\alpha=0.20$, the largest miscoverage we observe anywhere). Set sizes are below one passage on average and verdict accuracy at $\alpha=0.10$ is \FvAlphaTenDec. Unlike VitaminC, FEVER's label mix is balanced ($\FvDecSupp$ \supp{}, $\FvDecRef$ \refute{}, $\FvDecNei$ NEI of $\FvDecN$), so it is the setting in which verdict quality is actually measurable. \Cref{tab:fever-verdict} reports it: the one-sided conformal baseline never emits \refute{} and collapses to $\FvVerdSuppoDec$ decision accuracy, whereas CES reaches $\FvVerdCesDec$, on par with top-1 NLI ($\FvVerdToponeDec$) while returning calibrated evidence on both sides instead of at most one passage. This is the comparison that \cref{tab:main}'s decision column cannot make.

Across five shuffled draws the direction is stable (support $\MsFCovMeanAlphaTen \pm \MsFCovStdAlphaTen$, refutation $\MsFRefCovMeanAlphaTen \pm \MsFRefCovStdAlphaTen$ at $\alpha=0.10$); shuffling reduces the support gap relative to the in-order draw, consistent with page-associated heterogeneity rather than failure of the conformal construction itself.

FEVER also tests abstention quality, which VitaminC structurally cannot (its grouped cases always carry gold evidence on some side): $519/1500$ test claims are gold-NEI, giving abstention precision \FvAbstPrec{} and recall \FvAbstRec{} at $\alpha=0.10$. In this FEVER derivative the passages paired with NEI claims are retrieved by an automated system rather than human-annotated, so abstention quality is measured against system-selected non-evidence. Applying the same grouping diagnostic to VitaminC shows only mild page-cluster variation (support coverage spans $[\GcPageMin, \GcPageMax]$ across $\GcPageNGroups$ page clusters at $\alpha=0.10$), largely consistent with sampling uncertainty.

\begin{table}[t]
\centering
\caption{Verdict quality on FEVER at $\alpha=0.10$, where the gold label mix is balanced ($\FvDecSupp$/$\FvDecRef$/$\FvDecNei$ \supp{}/\refute{}/NEI of $\FvDecN$). The one-sided conformal baseline cannot emit \refute{} and loses over $20$ points of decision accuracy; CES matches top-1 NLI while also calibrating both evidence sides.}
\label{tab:fever-verdict}
\setlength{\tabcolsep}{4pt}
\footnotesize
\begin{tabular}{lcccc}
\toprule
Method & Supp.\ Cov. & Ref.\ Cov. & Dec.\ Acc. & Abst.\ Prec. \\
\midrule
Top-1 NLI        & \FvVerdToponeCov & \FvVerdToponeRefCov & \FvVerdToponeDec & \FvVerdToponeAbstPrec \\
Fixed thr.       & \FvVerdFixedCov  & \FvVerdFixedRefCov  & \FvVerdFixedDec  & \FvVerdFixedAbstPrec \\
Conf.\ supp-only & \FvVerdSuppoCov  & \FvVerdSuppoRefCov  & \FvVerdSuppoDec  & \FvVerdSuppoAbstPrec \\
\textbf{CES (ours)} & \textbf{\FvVerdCesCov} & \textbf{\FvVerdCesRefCov} & \FvVerdCesDec & \FvVerdCesAbstPrec \\
\bottomrule
\end{tabular}
\end{table}

\textbf{Group-stratified recalibration.} As an exploratory diagnostic repair, chosen after observing the undercoverage, we cluster claims by the first letter of the primary Wikipedia page name, compute group-specific $\tau^+_\alpha$ when at least \FvStratMinCal{} calibration examples are available, and otherwise fall back to the global threshold. \Cref{tab:fever-stratified} shows that this recovers near-nominal support coverage at $\alpha=0.05$ and substantially reduces the support-side gap at $\alpha=0.10$ and $0.20$, consistent with page-associated heterogeneity; the larger refutation-side gap is not addressed by this support-only stratification.

\begin{table}[t]
\centering
\caption{Group-stratified split-conformal on FEVER. ``Global'' uses one $\tau^+_\alpha$ across all claims; ``Stratified'' uses per-group $\tau$ on first-letter Wikipedia page-name clusters with $\ge\!\FvStratMinCal$ calibration examples (otherwise global fallback). ``Recovery'' is the ratio of the global to the stratified deviation from nominal.}
\label{tab:fever-stratified}
\setlength{\tabcolsep}{4pt}
\footnotesize
\begin{tabular}{lccc}
\toprule
$\alpha$ & Global supp.\ cov.\ (dev.) & Stratified supp.\ cov.\ (dev.) & Recovery \\
\midrule
$0.05$ & $\FvStratGlobalCovFive$ ($-3.6$ pp) & $\FvStratCovFive$ ($-\FvStratDevFive$ pp) & $9\times$ \\
$0.10$ & $\FvStratGlobalCovTen$ ($-6.7$ pp) & $\FvStratCovTen$ ($-\FvStratDevTen$ pp) & $2.1\times$ \\
$0.20$ & $\FvStratGlobalCovTwenty$ ($-5.3$ pp) & $\FvStratCovTwenty$ ($-\FvStratDevTwenty$ pp) & $3.5\times$ \\
\bottomrule
\end{tabular}
\end{table}

\subsection{Robustness to NLI Choice}\label{sec:nliswap}
Because the default NLI checkpoint includes FEVER in its fine-tuning mixture, we repeat the main experiment with \texttt{roberta-large-mnli}, fine-tuned on MNLI rather than FEVER. Support coverage remains within $1.4$--$3.1$ pp of nominal (\NliCovAlphaFive, \NliCovAlphaTen, \NliCovAlphaTwenty); the small deficit persists across five draws (\MsRCovMeanAlphaTen $\pm$ \MsRCovStdAlphaTen{} at $\alpha=0.10$). The weaker scorer produces larger sets (\NliSizeAlphaTen{} vs. \CesSizeAlphaTen{}) and lower verdict accuracy (\NliDecAlphaTen{} vs. \CesDecAlphaTen{}), illustrating that calibration stability does not imply verdict quality.

A dominance-margin sweep makes the separation sharper. With the default scorer, increasing $\delta$ raises accuracy among decided claims from \DlDebSelDZero{} to \DlDebSelDFifty{} while increasing abstention. Under the weak scorer, selective accuracy instead falls from \DlRobSelDZero{} to \DlRobSelDFifty{}; $|\Delta|$ is negatively associated with correctness (point-biserial correlation \CorrRob{}, versus \CorrDeb{} for the default scorer). A weak scorer therefore cannot be repaired simply by demanding a larger margin.

\subsection{Cross-Dataset Transferability}\label{sec:transfer}
At $\alpha=0.10$, within-dataset VitaminC calibration reaches the $0.90$ target (\XdCesCovVV), whereas FEVER$\to$FEVER remains below target (\XdCesCovFF), consistent with \cref{sec:fever}. Off-diagonal transfer drifts substantially: FEVER$\to$VitaminC gives \XdCesCovFV{} and VitaminC$\to$FEVER gives \XdCesCovVF{}. The tuned fixed-threshold baseline drifts similarly because the score distributions differ sharply (VitaminC $\tau^+_\alpha=0.918$ vs. FEVER $0.405$). CES therefore still requires a labeled calibration draw from the deployment distribution; its advantage is that recalibration is one quantile computation rather than a threshold grid search or test-set tuning.

\subsection{Ablation: Distractor Robustness}\label{sec:distract}
Randomly sampled negatives may be easier than real retrieval errors, so we also mine top-BM25 passages from other case ids, excluding near-duplicates of gold evidence. Because distractors add candidates without removing gold evidence, coverage is invariant by construction; the relevant effect is set quality. At six distractors, BM25 adds only about $0.2$--$0.3$ passages relative to random negatives, with precision within $2$ pp and verdict accuracy within $3$ pp; runs with $3$ and $10$ distractors show the same pattern. Some mined passages may actually bear on the claim despite lacking a gold label, so BM25 precision is mildly underestimated.

\begin{table}[t]
\centering
\caption{Distractor robustness at $\alpha=0.10$. Coverage is invariant by construction; BM25 hard negatives have only a modest effect on set quality.}
\label{tab:distract}
\setlength{\tabcolsep}{4pt}
\begin{tabular}{lccccc}
\toprule
Distractors & $n$ & Cov. & Size & Prec. & Abst. \\
\midrule
none   & 0 & \AblateCovDZero & \AblateSizeDZero & \AblatePrecDZero & \AblateAbstDZero \\
random & 6 & \AblateCovDSix  & \AblateSizeDSix  & \AblatePrecDSix  & \AblateAbstDSix \\
BM25   & 6 & \BmCovDSix      & \BmSizeDSix      & \BmPrecDSix      & \BmAbstDSix \\
\bottomrule
\end{tabular}
\end{table}

\subsection{End-to-End Pilot}\label{sec:e2e}
We run a small live-pipeline check on $60$ VitaminC validation claims rephrased as questions: BM25 retrieves top-$8$ passages from a $15{,}000$-passage corpus, Qwen2.5-7B-Instruct generates answers and decomposes them into \EpN{} atomic claims, and the VitaminC-calibrated CES layer is applied unchanged at $\alpha=0.10$. An LLM annotator independent of the NLI scorer labels every (claim, retrieved passage) pair; the annotation protocol is specified in \cref{sec:reproducibility} and released with the code.

The pilot exposes substantial front-end shift. BM25 retrieves no supporting passage for $176$ atomic claims, and the generator emits many meta-claims about missing context (\EpMeta{} of \EpN), so \cref{prop:coverage} does not formally apply. On the small evidence-eligible subsets, observed inclusion is nevertheless high: support succeeds on $17/\EpSuppElig$ claims (\EpSuppCov; Wilson 95\% CI $[0.74,0.99]$) and refutation on $35/\EpRefElig$ (\EpRefCov). Verdict transfer is poor: a fraction $\EpFalseSupp$ of the claims with no labeled support are still labeled \supp{}, and none of the \EpFree{} evidence-free claims trigger abstention. The pilot therefore reinforces the separation between calibrated evidence inclusion and uncalibrated retrieval/verdict quality.

\section{Discussion}
The main result is a separation between two properties that RAG evaluations often conflate. Under exchangeability, CES gives marginal finite-sample inclusion coverage for each evidence side and a Bonferroni joint bound; empirically, VitaminC tracks nominal closely and the joint bound is only mildly conservative. CES reaches essentially the same coverage--size frontier as a swept fixed threshold, but with a prespecified pointwise guarantee and without test exposure. On contrastive claims, one-sided conformal methods necessarily suppress refuting evidence, whereas CES can surface both sides.

The robustness studies explain when that guarantee is useful. Coverage remains comparatively stable across NLI scorers, random/BM25 distractors, and independent seeds, but degrades under FEVER heterogeneity and cross-dataset transfer; support-side stratification reduces the diagnosed FEVER support gap, while deployment-distribution recalibration is required under broader transfer. Verdict accuracy, precision, and dominance margins instead track scorer quality, and a weak scorer's largest margins can be its least reliable. In short: protect evidence coverage by monitoring calibration/deployment exchangeability, protect the verdict by improving or calibrating the scorer itself, and report the two regimes separately, because a single headline number conflates a property of the framework with a property of the plugged-in model.

\paragraph{Limitations.} Coverage is conditional on relevant evidence being present in the retrieved pool and is not a factual-correctness guarantee. VitaminC is deliberately contrastive and contains no gold-NEI claims, so it overstates how often real retrieval supplies evidence on both sides; FEVER and the live pilot expose this limitation. The FEVER grouping is dataset-specific, the NLI scorer inherits its training biases, and retrieval, generation, and decomposition remain outside the formal guarantee. CES also scores passages independently and does not explicitly model claims that require jointly sufficient multi-passage evidence. Extending calibration to end-to-end answers and to the verdict layer is future work.

\section{Reproducibility}\label{sec:reproducibility}
Experiments run on one Apple M5 Max workstation. The default scorer is \texttt{Mor\-itz\-Lau\-rer/De\-BER\-Ta-v3-large-mnli-fever-anli-ling-wanli}; VitaminC is \texttt{tals/vitaminc}. Calibration uses the ceiling split-conformal quantile with side-specific $n_\pm$. VitaminC calibration and evaluation use train/validation claims with disjoint \texttt{case\_id}s ($0/18{,}836$ validation case ids appear in train), preventing revisions of the same paragraph from crossing the split.

Pilot retrieval uses \texttt{rank\_bm25} over a deduplicated $15{,}000$-passage VitaminC corpus with lowercase alphanumeric tokenization; hard distractors use the same tokenizer and exclude the claim's case id and near-duplicates. Generation/decomposition use \texttt{Qwen2.5-7B-Instruct} (4-bit, MLX); pilot evidence labels are adjudicated by Claude Fable 5 (Anthropic, \texttt{claude-fable-5}), independent of the NLI scorer. Software uses Python 3.12, PyTorch 2.12, Transformers 5.12, and \texttt{datasets} 5.0; code, prompts, pilot labels, and the result files needed to reproduce the reported experiments are available at \url{https://github.com/BenRossJenkins/conformal-evidence-sets}.

Headline CES configurations, FEVER, and the NLI swap are replicated over \MsNSeeds{} independent draws using seeds $\{11,22,33,44,55\}$. Across all \MsNCells{} coverage cells, the largest seed s.d. is \MsMaxCovStd{} (\MsVMaxCovStd{} on VitaminC); VitaminC seed means track nominal at each level. Verdict accuracy varies more (\MsDecMean $\pm$ \MsDecStd{} at $\alpha=0.10$), with the reference single-seed point about two standard deviations above the mean, so verdict point estimates should be read accordingly.

\section{Conclusion}
Conformal Evidence Sets provide a two-sided contrastive grounding guarantee: calibrated support and refutation sets with finite-sample inclusion coverage on each side and a joint $1-2\alpha$ bound. On VitaminC, CES tracks nominal coverage and reaches the same empirical coverage--size frontier as a tuned fixed threshold while additionally providing the formal two-sided guarantee. Universal calibration, FEVER stress tests, scorer swaps, hard negatives, multi-seed replication, and a live pipeline pilot delineate the boundary of that guarantee.

The resulting deployment principle is simple: under exchangeability, \emph{evidence inclusion is calibrated; verdicts are earned by the scorer}. Distribution shift requires recalibration, retrieval failure invalidates the evidence premise, and an unreliable scorer cannot be rescued merely by increasing a dominance threshold. Future work should calibrate the verdict itself and extend the analysis to fully end-to-end long-form RAG.

\IEEEtriggeratref{12}
\bibliographystyle{IEEEtran}
\bibliography{ictai_references}

\end{document}